\documentclass{article}

\usepackage[preprint]{neurips_2024}

\usepackage[utf8]{inputenc} % allow utf-8 input
\usepackage[T1]{fontenc}    % use 8-bit T1 fonts
\usepackage{hyperref}       % hyperlinks
\usepackage{url}            % simple URL typesetting
\usepackage{booktabs}       % professional-quality tables
\usepackage{amsfonts}       % blackboard math symbols
\usepackage{nicefrac}       % compact symbols for 1/2, etc.
\usepackage{microtype}      % microtypography
\usepackage{xcolor}         % colors

\usepackage{todonotes}      % notes
\usepackage{amsmath}        % math symbols
\usepackage{mathtools}      % more math symbols
\usepackage{dsfont}         % even more math symbols
\usepackage{amsthm}         % theorem notation
\usepackage{cleveref}       % reference annotations

\newcommand{\dE}{\mathop{\mathbb{E}}}

\newtheorem{definition}{Definition}[section]
\newtheorem{lemma}{Lemma}[section]
\newtheorem{theorem}{Theorem}
\newtheorem{corollary}{Corollary}

\title{
On the ERM Principle in Meta-Learning
}

\author{
  Yannay Alon \\
  Technion \\
  \texttt{yannay.alon@campus.technion.ac.il} \\
  \And
  Steve Hanneke \\
  Purdue University \\
  \texttt{steve.hanneke@gmail.com}
  \And
  Shay Moran \\
  Technion \& Google Research \\
  \texttt{smoran@technion.ac.il} \\
  \And
  Uri Shalit \\
  Tel Aviv University \& Technion \\
  \texttt{urishalit@tauex.tau.ac.il} \\
}

\begin{document}

\maketitle

\begin{abstract}
Classic supervised learning involves algorithms trained on $ n $ labeled examples to produce a hypothesis $ h \in \mathcal{H} $ aimed at performing well on unseen examples. Meta-learning extends this by training across $ n $ tasks, with $ m $ examples per task, producing a hypothesis class $ \mathcal{H} $ within some meta-class $ \mathbb{H} $. This setting applies to many modern problems such as in-context learning, hypernetworks, and learning-to-learn. A common method for evaluating the performance of supervised learning algorithms is through their learning curve, which depicts the expected error as a function of the number of training examples. In meta-learning, the learning curve becomes a two-dimensional ``learning surface,'' which evaluates the expected error on unseen domains for varying values of $ n $ (number of tasks) and $ m $ (number of training examples).

Our findings characterize the distribution-free learning surfaces of meta-Empirical Risk Minimizers when either $ m $ or $ n $ tend to infinity: we show that the number of tasks ($ n $) must increase inversely with the desired error. In contrast, we show that the number of examples ($ m $) exhibits very different behavior: it satisfies a dichotomy where every meta-class conforms to one of the following conditions: (i) either $ m $ must grow inversely with the error, or (ii) a \emph{finite} number of examples per task suffices for the error to vanish as $ n $ goes to infinity. This finding illustrates and characterizes cases in which a small number of examples per task is sufficient for successful learning. We further refine this for positive values of $ \varepsilon $ and identify for each $ \varepsilon $ how many examples per task are needed to achieve an error of $ \varepsilon $ in the limit as the number of tasks $ n $ goes to infinity. We achieve this by developing a necessary and sufficient condition for meta-learnability using a bounded number of examples per domain.
\end{abstract}

\section{Introduction}
In classical learning problems, we are given examples sampled from some unknown distribution and wish to find a hypothesis that fits new examples from the same distribution. This general framework is the basis of many machine-learning systems used nowadays, from complex NLP (\citet{chowdhary2020natural,torfi2020natural,khurana2023natural}) and multi-modal tasks (\citet{baltruvsaitis2018multimodal,wang2023large}) through protein folding (\citet{jumper2021highly}) to decision support in healthcare (\citet{johnson2016machine,awaysheh2019review}) and many more.

However, when we consider human learning, we see that humans have the incredible ability not just to learn from specific examples as in supervised learning, but also to learn from tasks, and generalize to new tasks (\citet{bion2023learning}). In recent years, machine learning has followed this path with the idea of meta-learning (\citet{hospedales2021meta,ruder2017overview,howard2018universal}).
Meta-learning involves not only learning from data within a specific domain but also leveraging data from related domains. This approach is particularly useful when it allows us to solve a group of problems more efficiently than solving each problem individually.
For example, consider the problem of transcribing voice messages from different individuals. Each person's voice, accent, and speaking style represent a distinct domain. While there may not be enough data to train a robust model for each individual separately, we can leverage the commonalities across these domains to improve overall efficiency and performance in transcribing all voice messages.

In meta-learning, we aim to find a hypothesis class (or more generally, an algorithm) based on previous tasks or domains, that can be efficiently adapted to new domains.
This general framework captures many training strategies in deep learning including fine-tuning, in-context learning, few-shot learning, and more.
For example, in the case of pre-training and fine-tuning, one starts by training a model over large and versatile datasets to achieve an initial model. This stage induces a hypothesis class represented by all models that can be fine-tuned from this pre-trained model.
Next, when one tackles a specific domain, the pre-trained model is updated using some examples from the new domain which is equivalent to searching for a good hypothesis in the induced hypothesis class.

Following ideas from \citet{baxter2000model}, we assume that domains are sampled i.i.d from an (unknown) meta-distribution. We consider a meta-hypothesis family $\mathbb{H}$, which is a set of hypothesis classes.
We use data from the training domains in order to produce a hypothesis class $\mathcal{H}$ (not necessarily from $\mathbb{H}$) which in turn we use to find a good hypothesis for the test domain, based on examples from this domain.
The number of examples from the new domain needed for learning a good hypothesis from the produced class $\mathcal{H}$ is well studied under the distribution-free context in the PAC (Probably Approximately Correct) model. We thus focus on the first stage of meta-learning - how to learn a good hypothesis class based on examples from multiple training domains.

\subsection{Meta-learnability}
A common definition for meta-learnability (following \citet{baxter2000model}),  is that $\mathbb{H}$ is meta-learnable if there exists a meta-algorithm $\mathcal{A}$ such that for any desired error $\varepsilon > 0$, using large enough (yet finite) values for $n$, the number of domains, and $m$, the number of examples per domain, the meta-algorithm $\mathcal{A}$ produces a hypothesis class $\mathcal{H} \in \mathbb{H}$ that can later be fitted a new domain drawn from the same meta-distribution, using a finite sample from this domain, achieving an error of at most $\varepsilon$ in expectation.

In this paper, we focus on the case where the number of classes is finite (yet each class can be of infinite size). We show that without such a restriction, there are problems whose learnability cannot be determined. We further focus on meta-ERM algorithms.
Given the fundamental role of the ERM principle within the PAC model, and building on previous works in meta-learning that focus on deriving bounds for ERMs, we find meta-ERMs to be a natural path to initiate this investigation.

\subsection{Learning surface}
To study the performance of meta-ERM algorithms, we generalize the notion of a learning curve to a learning surface.
In classical learning, the learning curve describes the performance of an algorithm using the error rate as a function of the number of examples given.
In meta-learning, we have a learning surface that measures the performance of the meta-algorithm using the error rate as a function of both the number of domains $n$ and the number of examples per domain $m$. Several questions naturally arise for meta-learning:
Does the number of domains have to tend to infinity for the error to vanish?
Does the number of examples per domain have to tend to infinity?

In the PAC model, it is known that (except for trivial cases) the number of examples must grow indefinitely for the error to vanish.
We show that similarly, the number of domains must tend to infinity (except for trivial meta-hypothesis families).
Interestingly, previous results by \citet{aliakbarpour2024metalearning} show that for the number of examples per domain, there are cases when a finite number is sufficient for the error to vanish, and prove a sufficient condition for when this happens.
In this work, we improve upon this condition and give a necessary and sufficient condition characterizing which meta-hypothesis families can be learned using meta-ERMs with a bounded number of examples per domain.
Furthermore, we give an exact quantitative characterization of the number of examples per domain required to achieve any error $\varepsilon \geq 0$ for any meta-hypothesis family by meta-ERMs. 

\section{Problem setup}
\subsection{Meta-learning and the PAC model}
In the PAC model, a classification problem is defined by a distribution (or a domain) $D$ over the set of labeled examples $\mathcal{X} \times \mathcal{Y}$.
For binary classification, the setting in interest in this paper, we set $\mathcal{Y} = \{0, 1\}$.
The distribution $D$ is unknown, the learning algorithm $\mathcal{A}$ gets $n$ i.i.d examples from the domain: $S \sim D^n$ known as the training set.
The learning algorithm is a mapping from the training set to a classifier, $\mathcal{A}(S) \colon \mathcal{X} \to \mathcal{Y}$.
The task of the algorithm is to minimize the error of mislabeling an example drawn from the same domain:
$
    L_D(h) \coloneqq \Pr_{(x, y) \sim D} \left( h(x) \neq y \right)
$.
We say that $\mathcal{A}$ is an ERM algorithm if its output is a hypothesis consistent with its training set. That is, for $S = \{(x_i, y_i) \}_{i=1}^{n}$ the output $h = \mathcal{A}(S)$ satisfies for all $i \in [n]$: $h(x_i) = y_i$.
In the PAC model, we define a hypothesis class $\mathcal{H} \subseteq \mathcal{Y}^\mathcal{X}$, a set of classifiers. A domain $D$ is realizable with respect to $\mathcal{H}$ if it satisfies $\inf_{h \in \mathcal{H}} L_D(h) = 0$.

To generalize the PAC model for learning using multiple domains we follow the work of \citet{baxter2000model}.
In this framework, we assume a shared (unknown) meta-distribution over the possible domains $Q$, with the data our learning algorithm gets, sampled in two steps:
\begin{itemize}
    \item Sample $n$ i.i.d domains $D_1, \dots, D_n \sim Q^n$
    \item For each domain $i$, sample $m$ i.i.d examples $S_i = \{(x_{i, 1}, y_{i, 1}), \dots, (x_{i, m}, y_{i, m}) \} \sim D_i^m$
\end{itemize}
The final data is
$S = \left\{ \left\{ (x_{i, j}, y_{i, j}) \right\}_{j=1}^m \right\}_{i=1}^n$.
We denote this sampling process as $S \sim Q^{(n, m)}$.
The (meta-)algorithm $\mathcal{A}$ gets the training set $S$ and outputs an entire hypothesis class $\mathcal{H} \subseteq \mathcal{Y}^\mathcal{X}$ with the objective of minimizing the error
$L_Q(\mathcal{H}) \coloneqq \dE_{D \sim Q} \inf_{h \in \mathcal{H}} L_D(h)$.
Intuitively, we ask the algorithm to output a hypothesis class that contains a good hypothesis for future domains.
Later, when we want to learn a hypothesis for a specific domain, we search for a hypothesis in this class that best fits the new domain.
For the second stage to be feasible using a finite number of examples, we demand the output hypothesis class to have a finite VC dimension.

Like the hypothesis class $\mathcal{H}$ in the PAC model, we define a meta-hypothesis family $\mathbb{H}$ with each $\mathcal{H} \in \mathbb{H}$ being a hypothesis class.
The realizability property now becomes
$
    \inf_{\mathcal{H} \in \mathbb{H}} L_Q(\mathcal{H}) = 0
$. 
Similar to the development of the PAC model, we focus on the realizable settings, where the unknown meta-distribution $Q$ is realizable, and leave the extension to the agnostic case for future work.
We denote the set of all realizable meta-distributions as
$
    \text{RE}(\mathbb{H}) = \left\{ Q \;\middle|\; \inf_{\mathcal{H} \in \mathbb{H}} L_Q(\mathcal{H}) = 0 \right\}
$

As discussed, we require each class in our meta-hypothesis family to be a VC class:
\begin{definition}[VC family]
    A meta-hypothesis family $\mathbb{H}$ is a \underline{VC family} if there is $d \in \mathbb{N}$ such that $\forall \mathcal{H} \in \mathbb{H}$, $\text{VC}(\mathcal{H}) \leq d$.
    The \underline{VC of the meta-hypothesis family} is then $\text{VC}(\mathbb{H}) = \sup_{\mathcal{H} \in \mathbb{H}} \text{VC}(\mathcal{H})$.
\end{definition}
In this paper we study finite VC families, meaning there are finitely many classes in $\mathbb{H}$.
Note that a finite family can contain classes of infinite size.
Relaxing the finite assumption produces non-trivial subtleties which are beyond the scope of this paper: As we will later see, we can have meta-hypothesis families whose learnability cannot be determined (using the standard set theory axioms) when removing this assumption, even when $\text{VC}(\mathbb{H}) = 1$.

The meta-ERM algorithm in the meta-learning setting is a learning algorithm that outputs a hypothesis class $\mathcal{H}$ consistent with the training set:
For each training domain $i \in [n]$ there is a hypothesis $h_i \in \mathcal{H}$ such that for each example in the domain $j \in [m]$ we have $h_i(x_{i, j}) = y_{i, j}$.
As discussed in the introduction, we will focus on proper meta-ERM algorithms.

\subsection{Learning surface}
The learning surface of a specific algorithm over a specific meta-distribution is a function that assigns for each pair of $(n, m)$, the number of domains and the number of examples per domain respectively, the expected error of the algorithm over the distribution.

In the theory of PAC and beyond, the worst-case learning curve of an arbitrary ERM is an object of great interest. This object bounds the error an ERM might suffer from.
We extend this notion to the meta-learning setting.
\begin{definition}[ERM learning surface]
    Let $\mathbb{H}$ be a meta-hypothesis family.
    We define the \underline{ERM learning surface over $\mathbb{H}$} to be the function $\varepsilon^\text{ERM} \colon \mathbb{N} \times \mathbb{N} \to [0, 1]$ defined by:
    \[
        \varepsilon^\text{ERM}(n, m) = \sup_{Q \in \text{RE}(\mathbb{H})} \dE_{S \sim Q^{(n, m)}} \sup_{\mathcal{H} : L_S(\mathcal{H}) = 0} L_Q(\mathcal{H})
    \]
\end{definition}
Note the $\varepsilon^\text{ERM}$ chooses the highest error among classes consistent with the data. Since any ERM must output a consistent class, this bounds from above the worst-case behavior of ERMs.
To prove an upper bound for this learning surface, one must bound the error for all ERM algorithms uniformly for all realizable meta-distributions.
For a lower bound it is enough to find a single ERM with high error for some realizable meta-distribution.

In the PAC model, the fundamental theorem (\citet{vapnik1974theory}, \citet{ehrenfeucht1989general}) and the results from \citet{hanneke2016refined} show that there are three possible shapes for learning curves of ERMs using $n$ labeled examples: $\Theta_\mathbb{H} \left( \frac{1}{n} \right)$, $\Theta_\mathbb{H} \left( \frac{\log n}{n} \right)$ or $\Theta(1)$ depending on the finiteness of the star number \citet{hanneke2015minimax} and of the VC-dimension, where $\Theta_\mathbb{H}$ hides constants that may depend only on $\mathbb{H}$.

\subsection{Main results}
We start with the task of upper bounding the ERM learning surface $\varepsilon^\text{ERM}$ for finite VC families:
\begin{theorem}[ERM learning surface upper bound] \label{Theorem: General upper bound}
    Let $\mathbb{H}$ be a finite VC meta-hypothesis family. Then,
    \[
        \varepsilon^\text{ERM}(n, m) = O_\mathbb{H}\left( \frac{1}{n} + \frac{\log m}{m} \right)
    \]
    Where $O_\mathbb{H}$ hides constants that may depend on $\mathbb{H}$ only.
\end{theorem}
\begin{proof}[Proof idea]
    We bound the error of any ERM algorithm using two error causes.
    The first is where the sampled domains fail to approximate the meta-distribution, and the second is when the sampled examples fail to approximate the domains.

    The first term is bounded as a function of the number of domains, and the second is bounded using the number of examples per domain.
\end{proof}
The full proof is available in \Cref{Proof: general upper bound}

This bound suggests that as the number of domains and the number of examples per domain tend to infinity, the error vanishes and gives an upper bound on the required rate of each resource. We ask whether this bound can be met with a corresponding lower bound.
To answer this question, we separate between the two resources ($n$ and $m$) and define the projection of the learning surface for a single resource:
\begin{definition}[Learning surface projection]
    Let $\mathbb{H}$ be a meta-hypothesis family and denote its corresponding ERM learning surface $\varepsilon^\text{ERM}$.
    The \underline{projections of $\varepsilon^\text{ERM}$} on each resource are:
    \begin{align*}
        &\varepsilon^\text{ERM}_\text{dom} (n) = \lim_{m \to \infty} \varepsilon^\text{ERM}(n, m)
        &\varepsilon^\text{ERM}_\text{exp} (m) = \lim_{n \to \infty} \varepsilon^\text{ERM}(n, m)
    \end{align*}
\end{definition}
The learning surface is monotonic for each variable and bounded. Hence the limits are well-defined.
From the result of \Cref{Theorem: General upper bound}, as each resource tends to infinity we have the upper bounds:
\begin{corollary} \label{Corollary: projections upper bound}
    Let $\mathbb{H}$ be a finite VC meta-hypothesis family. Then,
    \begin{align*}
        &\varepsilon^\text{ERM}_\text{dom}(n) = O_\mathbb{H} \left( \frac{1}{n} \right)
        &\varepsilon^\text{ERM}_\text{exp}(m) = O_\mathbb{H} \left( \frac{\log m}{m} \right).
    \end{align*}
\end{corollary}
Next, we define a simple notion of non-triviality for meta-hypothesis families.
\begin{definition}[Informal definition of non-trivial meta-hypothesis family]
    Let $\mathbb{H}$ be a meta-hypothesis family. We say that \underline{$\mathbb{H}$ is non-trivial} if: there exists an example $(x, y)$ which is realizable by more than a single class,
    and no class is dominated by another class.
\end{definition}
The formal definitions for the two conditions are given in \Cref{Assumption: weak non-separability}, \Cref{Assumption: no pairwise domination}.
Intuitively, a meta-hypothesis family that does not satisfy the first condition can always be learned perfectly using only a single example from a single domain.
The second condition demands the meta-hypothesis family not contain any redundant classes that should never be produced.

The following parameter defines a sense of complexity for the meta-hypothesis family. For a more complete definition, see \Cref{Section: number of examples per domain}.
\begin{definition}[$\varepsilon$ dual Helly number] \label{definition: epsilon dual Helly number}
     Let $\mathbb{H}$ be a meta-hypothesis family.
     For any class $\mathcal{H} \in \mathbb{H}$ and $\varepsilon \geq 0$ we define the \underline{$\varepsilon$ dual Helly number of $\mathcal{H}$ with respect to $\mathbb{H}$}, denoted as $m_{\mathcal{H} \mid \mathbb{H}}(\varepsilon)$ to be the smallest integer $m$ such that any set $S \subseteq \mathcal{X} \times \mathcal{Y}$ which is realizable by $\mathbb{H}$ but is $\varepsilon$-non-realizable by $\mathcal{H}$ has a non-realizable subset $S' \subseteq S$ with $\lvert S' \rvert \leq m$.

     We further define the \underline{$\varepsilon$ dual Helly number of $\mathbb{H}$} as $m_\mathbb{H}(\varepsilon) = \sup_{\mathcal{H} \in \mathbb{H}} m_{\mathcal{H} \mid \mathbb{H}}(\varepsilon)$.
\end{definition}
In the special case where $\varepsilon = 0$, we recover the dual Helly number from \citet{bousquet2020proper} which is inspired by Helly's theorem, a fundamental result about convex sets.
See \Cref{Definition: dual Helly number} for more discussion about related dimensions.
Intuitively, this number represents the minimal amount of examples we need to ensure a realizable set hints the hypothesis class fits the data.
Using less than $m_\mathbb{H}(\varepsilon)$, any set that seems realizable does not tell us anything on whether the entire domain is $\varepsilon$-realizable or not.

Using these definitions we can complement the upper bounds with the lower bounds:
\begin{theorem}[Learning surface's projections lower bound] \label{Theorem: projections lower bound}
    For any non-trivial finite VC meta-hypothesis family, $\mathbb{H}$:
    \begin{enumerate}
        \item Number of domains must go to infinity: $\varepsilon^\text{ERM}_\text{dom}(n) = \Omega_\mathbb{H} \left(\frac{1}{n} \right)$.
        \item Dichotomy for the number of examples per domain:
        \begin{itemize}
            \item If $m_\mathbb{H}(0) = \infty$, the number of examples must go to infinity: $\varepsilon^\text{ERM}_\text{exp}(m) = \Omega_\mathbb{H} \left(\frac{1}{m} \right)$.
            \item Otherwise, for any $\varepsilon \geq 0$, we can achieve $\varepsilon$ error using a finite number of examples.
            Specifically, to achieve zero error; for all $m \geq m_\mathbb{H}(0)$ we have
            $\varepsilon^\text{ERM}_\text{exp}(m) = 0$ .
            More generally,
            $m \geq m_\mathbb{H}(\varepsilon) \iff \varepsilon^\text{ERM}_\text{exp}(m) \leq \varepsilon$.
            % Specifically,  
            % $m \geq m_\mathbb{H}(\varepsilon) \iff \varepsilon^\text{ERM}_\text{exp}(m) \leq \varepsilon$.
            % Importantly, $\varepsilon^\text{ERM}_\text{exp}(m) = 0$ for all $m \geq m_\mathbb{H}(0)$
        \end{itemize}
    \end{enumerate}
\end{theorem}
The proof for the number of domains is given in \Cref{Section: number of domains} and for the number of examples per domain in \Cref{Section: number of examples per domain}.

This result shows that in some non-trivial cases, a few examples per domain suffice for perfect meta-learning (as $n \to \infty$).
For example, consider some high-dimensional space $\mathcal{X} = \mathbb{R}^\ell$. For any set of vectors $V \subseteq \mathbb{R}^\ell$, denote by $\mathcal{H}_{V} = \left\{h_{w, b}(x) = \mathds{1}(w^T x \geq b) \;\middle|\; w \in V, b \in \mathbb{R} \right\}$ the class of all half-spaces whose normal is in $V$.
Let $\mathbb{H} = \left\{ \mathcal{H}_V \;\middle|\; V \in \mathcal{V} \right\}$ for some set finite set $\mathcal{V}$ with $\lvert V \rvert \leq d \ll \ell$ for all $V \in \mathcal{V}$.
Then we have $\text{VC}(\mathbb{H}) \leq d + 1$ and $m_\mathbb{H}(0) \leq d + 2$
\footnote{For the dual Helly of half-spaces, see proposition 2.8 in \citet{braverman2019convex} and the remark afterward.}.
Therefore, we can meta-learn $\mathbb{H}$ and achieve zero error with only $d + 2$ examples per domain.
Note that this number of examples is far too small to learn a good hypothesis with error $\varepsilon < \frac{1}{2}$ for each domain even for a specific $\mathcal{H}_V$.

Combined with the result of \Cref{Corollary: projections upper bound} we have a tight bound for the number of domains. Furthermore, this lower bound applies to all proper algorithms, not only meta-ERMs. This shows that any meta-ERM achieves the optimal sample complexity for the number of domains.

For the number of examples per domain, this result characterizes exactly which meta-hypothesis classes can achieve zero error using a finite number of examples per domain.
For such families, we know exactly how many examples per domain are needed to achieve $\varepsilon$ error, for any $\varepsilon \geq 0$

One might have expected that we could have families with arbitrarily fast rates regarding the number of examples per domain. However, this result shows a gap between the possible rates.
All families with unbounded dual Helly can only hope for a rate of $\nicefrac{1}{m}$ and no faster (like $e^{-m}$ or even $\frac{1}{m \log m}$).
For these families, we remain with a gap between the upper and lower bound.
As we will see, for some meta-hypothesis families the lower bound of $\frac{1}{m}$ is achievable while others require the upper bound of $\frac{\log m}{m}$, and rates in between are also possible.
% The exact rate is strongly connected with the concept of $\varepsilon$-nets, which is a well-studied object in computational geometry.

The case of finite examples per domain has been studied by \citet{aliakbarpour2024metalearning} where a sufficient condition for finite examples was proven.
Our result refines the parameter defined there, which allows for a sufficient and necessary condition for proper ERMs. Furthermore, our result generalizes the condition for finite examples for all error $\varepsilon \geq 0$.

\section{Related work}
The idea of meta-learning has been researched under many names including learning-to-learn, bias learning, transfer learning, multitask learning, domain generalization, hyperparameter learning, and more. In some cases, the exact term used suggests specific assumptions or modeling of the problem.
Since the work of \citet{baxter2000model} where the meta-learning framework we build upon was suggested and several upper bounds were obtained. Many following works have improved upon the upper bounds using different methods such as PAC-Bayes \citet{pmlr-v162-rezazadeh22a} (and citations within), information-theoretic \citet{chen2021generalization} and distribution-dependent bounds \citet{konobeev2021distribution, titsias2021information}.
Others used different notions of relatedness among the domains \citet{ben2003exploiting, tripuraneni2020theory, jose2021information, mahmud2009universal}, or focused on specific structures, from linear classification \citet{tripuraneni2021provable} up to deep networks \citet{galanti2016theoretical}.

The vast majority of previous work has been focused on finding upper bounds for the generalization performance of meta-learning. Some works complement these with lower bounds, either for special cases like linear representations \citet{tripuraneni2021provable, aliakbarpour2024metalearning} or under strong relatedness assumptions \citet{lucas2020theoretical}.
Other works show lower bounds for related problems like identifiability \citet{yang2011identifiability}.

The Empirical Risk Minimization (ERM) principle has been a long-time interest in the theory of machine learning. The works of \citet{vapnik1991principles, blumer1989learnability, alon1997scale} and many more investigated the role of ERM algorithms in binary classification.
The work of \citet{shalev2010learnability} has studied the limitations of ERMs in general learning settings.
Other works have studied the ERM principle in different settings like multi-class classification\citet{ben1992characterizations, daniely2011multiclass} and differential privacy \citet{wang2016learning}.
In this work, we study the ERM for the setting of multi-learning.
A line of works after \citet{pmlr-v70-finn17a} suggested gradient-based ERMs for meta-learning (\citet{nichol2018reptile, nichol2018first} and more).

The most related work to ours is \citet{aliakbarpour2024metalearning}, where a similar meta-learning model was researched and some sufficient conditions for learnability with a finite number of examples per domain were proven. In our work, we strengthen this result and prove a sufficient and necessary condition. Moreover, we generalize this result for a positive error rate, allowing for a trade-off between the sample complexity and an acceptable error.

\section{Analysis}
\subsection{Number of domains} \label{Section: number of domains}
For \Cref{Theorem: projections lower bound}
we assumed two conditions of non-triviality which we elaborate on here:
\begin{definition}[Weak non-separability] \label{Assumption: weak non-separability}
    Let $\mathbb{H}$ be a meta-hypothesis family. We say that the meta-hypothesis family $\mathbb{H}$ satisfies the \underline{weak non-separability assumption} if there exists an example $(x, y) \in \mathcal{X} \times \mathcal{Y}$ such that $\lvert \left\{ \mathcal{H} \in \mathbb{H} : \exists h \in \mathcal{H} \enspace h(x) = y \right\} \rvert > 1$.
\end{definition}
If $\mathbb{H}$ does not satisfy this assumption, then any example determines the only consistent hypothesis class. Consequently, a single example from a single domain is enough to achieve optimal error.

\begin{definition}[No pairwise domination]\label{Assumption: no pairwise domination}
    Let $\mathbb{H}$ be a meta-hypothesis family. We say the meta-hypothesis family $\mathbb{H}$ satisfies the \underline{no pairwise domination assumption} if for any two different hypothesis classes $\mathcal{H}, \mathcal{H}' \in \mathbb{H}$, there exists some meta-distribution $Q$ which is realizable by $\mathcal{H}'$ but not by $\mathcal{H}$, that is, $L_Q(\mathcal{H}) > 0$ and $L_Q(\mathcal{H}') = 0$.
\end{definition}
Suppose $\mathbb{H}$ does not satisfy this assumption. In that case, there is a redundant hypothesis class $\mathcal{H}' \in \mathbb{H}$ and a dominant one $\mathcal{H} \in \mathbb{H}$ such that $L_Q(\mathcal{H}') = 0 \implies L_Q(\mathcal{H}) = 0$, so $\mathcal{H}'$ can be removed.

A stronger version of the bound in \Cref{Theorem: projections lower bound} for the number of domains can now be stated:
\begin{theorem}[Learning surface lower bound with respect to $n$] \label{Theorem: lower bound number of domains}
    Let $\mathbb{H}$ be a meta-hypothesis family with $\lvert \mathbb{H} \rvert < \infty$.
    Assume $\mathbb{H}$ satisfies \Cref{Assumption: weak non-separability,Assumption: no pairwise domination}.
    Then, there exists a constant $c > 0$ depending only on $\mathbb{H}$ such that for any proper learning algorithm $\mathcal{A}$ and any $n, m \in \mathbb{N}$:
    \[
        \sup_{Q \in \text{RE}(\mathbb{H})}\dE_{S \sim Q^{(n, m)}} L_Q(\mathcal{A}(S)) \geq \frac{c}{n}
    \]
\end{theorem}
\begin{proof}[Proof idea]
    Using \Cref{Assumption: weak non-separability}, we construct an easy domain on which multiple hypothesis classes are realizable.
    For each algorithm, we define the distribution it induces over $\mathbb{H}$ when this domain is the only one sampled.
    For the mode of this distribution, using \Cref{Assumption: no pairwise domination}, there is a hard domain with high error.

    Finally, using a convex combination of the two distributions, we ensure that with a high enough probability, the training domains are all the same easy domain while the test domain is the hard one that achieves the desired error.
\end{proof}
The full proof is given in \Cref{Proof: lower bound number of domains}. We also give a more general result for infinite meta-hypothesis classes in \Cref{Theorem: lower bound number of domains - general case}.
Note that this result holds for all proper algorithms, not only for ERMs.
As we have shown for the upper bound, any meta-ERM achieves this optimal rate of $\frac{1}{n}$ with respect to the number of domains.

\subsection{Number of examples per domain} \label{Section: number of examples per domain}
For the number of examples per domain, we identify the dual Helly number \citet{bousquet2020proper} and some generalizations of it as an important characteristic in determining the behavior of the learning surface.
Intuitively, the dual Helly number is a parameter of the hypothesis class that measures how large a sample needs to be to allow the detection of a non-realizable domain.
More formally,
\begin{definition}[Dual Helly number] \label{Definition: dual Helly number}
    Let $\mathcal{H}$ be a hypothesis class.
    The \underline{dual Helly number of $\mathcal{H}$} is defined as the minimal size $m$ such that every non-realizable set has a subset of size at most $m$ that witnesses this non-realizability. Formally:
    \[
        DH(\mathcal{H}) \coloneqq
        \min \left\{ m :
            \forall S \subseteq \mathcal{X} \times Y
            \enspace L_S(\mathcal{H}) > 0
            \enspace \exists S' \subseteq S 
            \enspace \lvert S' \rvert \leq m
            \enspace L_{S'}(\mathcal{H}) > 0 
        \right\}
    \]
    In the case all $m \in \mathbb{N}$ do not satisfy the condition, we set $DH(\mathcal{H}) = \infty$.
\end{definition}
The dual Helly number, and some variants of it, has been shown to have a strong connection with the learnability in the PAC model for proper algorithms \citet{bousquet2020proper}, with proper learning in a distributed settings \citet{kane2019communication} where it was named coVC, and to allow meta-learning using a finite number of examples in the realizable scenario \citet{aliakbarpour2024metalearning} where it was named the Non-Realizable Certificate (NRC) complexity.
We generalize the notion of the dual Helly number for $\varepsilon$-non-realizable sets as follows:
\begin{definition}[$\varepsilon$ dual Helly number]
    Let $\mathcal{H}$ be a hypothesis class.
    For any $\varepsilon \in [0, 1]$ we define the \underline{$\varepsilon$ dual Helly number of $\mathcal{H}$} as:
    \[
        m_\mathcal{H}(\varepsilon) \coloneqq
        \min \left\{ m :
            \forall S \subseteq \mathcal{X} \times \mathcal{Y}
            \enspace L_S(\mathcal{H}) > \varepsilon
            \enspace \exists S' \subseteq S
            \enspace \lvert S' \rvert \leq m
            \enspace L_{S'}(\mathcal{H}) > 0
        \right\}
    \]
    That is the minimal size $m$ for which every $\varepsilon$ non-realizable set is witnessed by a subset of size at most $m$.
    If no $m \in \mathbb{H}$ satisfies the condition, we define $m_\mathcal{H}(\varepsilon) = \infty$.
\end{definition}
As we will see, this variant of the dual Helly number is strongly related to the learnability of a meta-hypothesis family for varying levels of acceptable error.
\Cref{definition: epsilon dual Helly number} is a variant of this definition where we only consider sets that are realizable by the meta-hypothesis family. This extra condition will allow us to prove a necessary and sufficient condition unlike the NRC in \citet{aliakbarpour2024metalearning} that only gives a sufficient condition.

Finally, we define the optimal error function of a meta-hypothesis family.
Intuitively, this can be considered as the inverse of the $\varepsilon$ dual Helly. If the dual Helly translates a desired error to the required number of examples, the optimal error is the minimal error guaranteed by this number of examples.
\begin{definition}[Optimal error function] \label{Definition: optimal error function}
    Let $\mathbb{H}$ be a meta-hypothesis family.
    We define the \underline{optimal error function of $\mathbb{H}$} as
    $
        \varepsilon_\mathbb{H}(m) \coloneqq \inf \left\{ \varepsilon \in [0, 1] \;\middle|\; m_\mathbb{H}(\varepsilon) \leq m \right\}
    $
\end{definition}
Using those definitions, we are ready to characterize the projection learning surface for $m$.
The following results show that the optimal error function is precisely the optimal rate at which the meta-hypothesis can be learned using an arbitrary ERM algorithm.

\begin{theorem}[Learning surface with respect to $m$ and dual Helly number] \label{Theorem: bound number of examples per domain}
    Let $\mathbb{H}$ be a meta-hypothesis family with $\lvert \mathbb{H} \rvert < \infty$ and $\text{VC}(\mathbb{H}) < \infty$.
    Then,
    \[
        \varepsilon^\text{ERM}_\text{exp}(m)
        = \varepsilon_\mathbb{H}(m)
    \]
\end{theorem}
\begin{proof}[Proof idea]
    We show two inequalities.
    For the upper bound, $\varepsilon^\text{ERM}_\text{exp}(m) \geq \varepsilon_\mathbb{H}(m)$, we note that for an ERM to make a mistake he must fail to identify a hypothesis class with an error greater than $\varepsilon_\mathbb{H}(m)$. This might happen when either the sampled domains fail to show the true error of the meta-distribution or the sampled examples fail to show the error of the domains.

    Instead of the two-stage sampling method, we can equivalently sample $S \sim Q^{(n, m)}$ in three steps.
    First sample the domains $\mathbf{D} \sim Q^n$,
    then sample an $\varepsilon$-net of size $\approx \frac{d + \log \frac{1}{\delta}}{\varepsilon^2}$ from each domain,
    and finally sample $m$ examples from each $\varepsilon$-net (with no repetition).

    Then, we bound the probability for each step that the sampled object fails to show an error in the previous step using conditioning.

    For the lower bound, $\varepsilon^\text{ERM}_\text{exp}(m) \leq \varepsilon_\mathbb{H}(m)$, we show that for all errors smaller than $\varepsilon_\mathbb{H}(m)$, using the contradiction of the $\varepsilon$ dual Helly number we can find a hard set of examples that is $\varepsilon$ non-realizable but cannot be witnessed by any sampled set using only $m$ examples per domain.

    Therefore, regardless of the number of domains, this hard sample would seem to be realizable, and an arbitrary ERM could consistently choose the wrong hypothesis class.
\end{proof}
The full proof is given in \Cref{Proof: bound number of examples per domain}.

Following this result, we see that the optimal error function $\varepsilon_\mathbb{H}$ separates between two cases.
When $\varepsilon_\mathbb{H}(m_0) = 0$ for some $m_0$, we do not need more than $m_0$ examples for the error to vanish. This happens exactly when the dual Helly number $m_\mathbb{H}(0)$ is finite.
Otherwise, we have $\varepsilon_\mathbb{H}(m) > 0$ for all $m$ and we must have $m$ tend to infinity for the error to go to zero.

The next result shows that the dual Helly number exactly characterizes the distinction between the need in an infinite number of examples per domain and the sufficiency of a finite one. Moreover, in the case of an infinite number of examples, one can only hope for a rate of $\frac{1}{m}$ and no faster.
\begin{lemma}[optimal error function and dual Helly number] \label{Lemma: Rates of dual Helly}
    Let $\mathbb{H}$ be a finite VC meta-hypothesis.
    \begin{itemize}
        \item If $m_\mathbb{H}(0) < \infty$, then for all $m \geq m_\mathbb{H}(0)$ we have $\varepsilon_\mathbb{H}(m) = 0$.
        \item Otherwise, if $m_\mathbb{H}(0) = \infty$,
        then for infinitely many $m \in \mathbb{N}$ we have $\varepsilon_\mathbb{H}(m) \geq \frac{1}{m + 1}$.
    \end{itemize}
\end{lemma}
The proof is given in \Cref{Proof: rates of dual Helly}.

Combining \Cref{Theorem: bound number of examples per domain} with \Cref{Lemma: Rates of dual Helly} gives \Cref{Theorem: projections lower bound} for the number of examples per domain.
For the case where $m_\mathbb{H}(0) = \infty$, we are left with a gap $\frac{1}{m} \leq \varepsilon^\text{ERM}_\text{exp}(m) \leq \frac{\log m}{m}$ (ignoring exact constants).

\subsection{Impossibility in the infinite case}
In this paper, we limited our attention to meta-hypothesis families with a finite number of classes.
In this section, we will see that relaxing this restriction raises subtleties where the notion of learnability itself becomes unstable, even when all hypothesis classes have VC dimension 1.

This will be done by showing the continuum hypothesis can be encoded as a learnability problem of a meta-hypothesis family.
Since the continuum hypothesis is independent of the Zermelo–Fraenkel set theory with the axiom of choice (ZFC), one cannot determine whether this meta-hypothesis family is learnable.

Given a finite subset $X \subseteq [0, 1]$, denote $\mathcal{H}_X = \left\{ \mathds{1}_{\{x\}} : x \in X \right\}$ the set of indicators over singletons in $X$.
Define the meta-hypothesis family $\mathbb{H}^* = \left\{ \mathcal{H}_X : X \subseteq [0, 1] \enspace \lvert X \rvert < \infty \right\}$.

\begin{theorem} \label{Theorem: independence result}
    The learnability of $\mathbb{H}^*$ is independent of the ZFC axioms.
\end{theorem}
\begin{proof}[Proof idea]
    We show that the learnability of $\mathbb{H}^*$ is an EMX problem, as defined in \citet{ben2019learnability}.
    In their work, they proved the EMX problem is independent of the ZFC axioms.
\end{proof}
This theorem indicates some limit for results that can be established for the proper meta-learning framework in its general form. Notably, no simple combinatorial dimension can fully characterize which meta-hypothesis families are learnable in the proper settings and which are not.
For a formal definition of such simple dimensions, see \citet{ben2019learnability}.
The proof of our result is given in \Cref{Proof: independence result}.

\section{Future work} \label{Section: future work}
There are several interesting directions for future research. To begin with, we assumed a fairly strong assumption on the structure of the meta-hypothesis family of being a finite family.
Even though this assumption cannot be relaxed completely due to the independence result of \Cref{Theorem: independence result}, there are possible options for limiting the meta-hypothesis family.
One approach would be assuming a finite dimension of the family, for example, a version of the Littlestone dimension or a VC dimension like the one suggested by \citet{aliakbarpour2024metalearning}.
Another approach would be assuming a parameterized family.

In the framework discussed here, we assumed the meta-algorithm to be proper, meaning the hypothesis class it produces must be from $\mathbb{H}$. Allowing the meta-algorithm to produce any hypothesis class would fail to capture the structure of the problem
\footnote{In this case, the meta-hypothesis family $\mathbb{H}$ is meta-learnable if and only if the union over $\mathbb{H}$ is a VC class.
An algorithm that outputs the union trivially meta-learn when the union is a VC class, and when the union is meta-learnable, applying the algorithm over a meta-distribution that outputs the same domain w.p. $1$ will PAC-learn the union.
However, this would result in poor sample complexity for the downstream tasks.}.
However, we can limit the meta-algorithm using a restricted notion of improperness. For example, we might restrict the algorithm to produce hypothesis classes whose VC dimension is $O(\text{VC}(\mathbb{H}))$. This will ensure the sample complexity for the downstream tasks will not suffer too much.

The focus of this paper was the ERM principle in the setting of meta-learning. There are several questions left open: Are ERMs optimal in this setting? Is there a gap in the asymptotic behavior between the worst and the best ERM?
For the number of domains, we have seen that ERMs are optimal among all proper meta-algorithms and all ERMs achieve the same asymptotic bound. For the number of examples per domain, these questions are still open.

In this work we only discuss the realizable case. Nevertheless, the agnostic case for the meta-learning setting is of great interest.
Some bounds can be easily proven using standard techniques from the PAC model combined with the analysis we developed here.

The meta-learning model we studied in this paper was in the distribution-free context.
However, other contexts with different motivations were developed over the years like the non-uniform setting \citet{benedek1994nonuniform} or the universal learning model \citet{bousquet2021theory} to name a few.

\section{Acknowledgements}
Shay Moran is a Robert J. Shillman Fellow; he acknowledges support by ISF grant 1225/20, by BSF grant 2018385, by Israel PBC-VATAT, by the Technion Center for Machine Learning and Intelligent Systems (MLIS), and by the European Union (ERC, GENERALIZATION, 101039692). Views and opinions expressed are however those of the author(s) only and do not necessarily reflect those of the European Union or the European Research Council Executive Agency. Neither the European Union nor the granting authority can be held responsible for them.

\bibliography{main.bib}

\newpage
\appendix

\section{Appendix}
\subsection{General upper bound}
\begin{proof}[Proof of \Cref{Theorem: General upper bound}] \label{Proof: general upper bound}
    We wish to bound uniformly for all $Q \in \text{RE}(\mathbb{H})$ the expected error of the worst ERM:
    \[
        \dE_{S \sim Q^{(n, m)}} \sup_{\mathcal{H} : L_S(\mathcal{H})} L_Q(\mathcal{H})
    \]
    We denote $\mathbb{H}[S] = \left\{ \mathcal{H} \in \mathbb{H} \;\middle|\; L_S(\mathcal{H}) = 0 \right\}$ for simplicity.

    Fix a sample $S \sim Q^{(n, m)}$ and denote its domains by $\mathbf{D} \in \mathcal{D}^n$.
    For any fixed $\mathcal{H} \in \mathbb{H}$ and $\varepsilon_0 > 0$ we have:
    \begin{align*}
        L_Q(\mathcal{H})
        &= \dE_{D \sim Q} L_D(\mathcal{H})
        = \int_0^1 \Pr_{D \sim Q} \left( L_D(\mathcal{H}) > \varepsilon \right) d \varepsilon \\
        &\leq \int_0^{\varepsilon_0} 1 d \varepsilon + \int_{\varepsilon_0}^1 \Pr_{D \sim Q} \left( L_D(\mathcal{H}) > \varepsilon_0 \right) d \varepsilon \\
        &= \varepsilon_0 + (1 - \varepsilon_0) \Pr_{D \sim Q} \left( L_D(\mathcal{H}) > \varepsilon_0 \right)
        \leq \varepsilon_0 + \Pr_{D \sim Q} \left( L_D(\mathcal{H}) > \varepsilon_0 \right)
    \end{align*}
    Denote $\lambda_Q(\mathcal{H}) = \Pr_{D \sim Q} \left( L_D(\mathcal{H}) > \varepsilon_0 \right)$. We divide into two cases:
    \begin{enumerate}
        \item If $\forall i \in [n]$, $L_{D_i}(\mathcal{H}) \leq \varepsilon_0$, then
        \[
            \lambda_Q(\mathcal{H})
            \leq \lambda_Q(\mathcal{H}) \cdot \mathds{1} \left( \forall i \in [n] : L_{D_i}(\mathcal{H}) \leq \varepsilon_0 \right)
        \]
        \item Otherwise, $\exists i \in [n]$ such that $L_{D_i}(\mathcal{H}) > \varepsilon_0$. Since $\lambda_Q(\mathcal{H}) \leq 1$, we have
        \[
            \lambda_Q(\mathcal{H})
            \leq \mathds{1} \left( \exists i \in [n] : L_{D_i}(\mathcal{H}) > \varepsilon_0 \right)
        \]
    \end{enumerate}
    In any case,
    \[
        \lambda_Q(\mathcal{H})
        \leq \lambda_Q(\mathcal{H}) \cdot \mathds{1} \left( \forall i \in [n] : L_{D_i}(\mathcal{H}) \leq \varepsilon_0 \right) + \mathds{1} \left( \exists i \in [n] : L_{D_i}(\mathcal{H}) > \varepsilon_0 \right)
    \]
    Taking the supremum over $\mathcal{H} \in \mathbb{H}[S]$ gives,
    \begin{align*}
        \sup_{\mathcal{H} \in \mathbb{H}[S]} \lambda_Q(\mathcal{H})
        &\leq \sup \left\{ \lambda_Q(\mathcal{H}) \;\middle|\; \mathcal{H} \in \mathbb{H}, \enspace \forall i \in [n] : L_{D_i}(\mathcal{H}) \leq \varepsilon_0 \right\} \\
        &+ \mathds{1} \left( \exists \mathcal{H} \in \mathbb{H}[S], \enspace \exists i \in [n] : L_{D_i}(\mathcal{H}) > \varepsilon_0 \right)
    \end{align*}
    We will now bound each term separately. For the first term, taking the expectation over $\mathbf{D} \sim Q^n$ using the formula for expectation of non-negative variable $\dE[X] = \int_{-\infty}^\infty \Pr(X \geq x) d x$,
    \begin{align*}
        \dE_{\mathbf{D} \sim Q^n} \sup &\left\{ \lambda_Q(\mathcal{H}) \;\middle|\; \mathcal{H} \in \mathbb{H}, \enspace \forall i \in [n] : L_{D_i}(\mathcal{H}) \leq \varepsilon_0 \right\} \\
        &= \int_0^1 \Pr_{\mathbf{D} \sim Q^n} \left( \sup \left\{ \lambda_Q(\mathcal{H}) \;\middle|\; \mathcal{H} \in \mathbb{H}, \enspace \forall i \in [n] : L_{D_i}(\mathcal{H}) \leq \varepsilon_0 \right\} \geq \gamma \right) d \gamma  \\
        &= \int_0^1 \Pr_{\mathbf{D} \sim Q^n} \left(\exists \mathcal{H} \in \mathbb{H} : \lambda_Q(\mathcal{H}) \geq \gamma \wedge \forall i \in [n] : L_{D_i}(\mathcal{H}) \leq \varepsilon_0  \right) d \gamma \\
        \intertext{Applying a union bound with the trivial bound for probabilities}
        &\leq \int_0^1 \min \left\{1, \sum_{\mathcal{H} \in \mathbb{H}} \Pr_{\mathbf{D} \sim Q^n} \left( \lambda_Q(\mathcal{H}) \geq \gamma \wedge \forall i \in [n] : L_{D_i}(\mathcal{H}) \leq \varepsilon_0  \right) \right\} d \gamma \\
        \intertext{Using conditional probabilities $\Pr(A \wedge B) = \Pr(B \mid A) \Pr(A) \leq \Pr(B \mid A)$}
        &\leq \int_0^1 \min \left\{1, \sum_{\mathcal{H} \in \mathbb{H}} \Pr_{\mathbf{D} \sim Q^n} \left( \forall i \in [n] : L_{D_i}(\mathcal{H}) \leq \varepsilon_0  \;\middle|\; \lambda_Q(\mathcal{H}) \geq \gamma \right) \right\} d \gamma \\
        \intertext{Since the domains are drawn i.i.d:}
        &= \int_0^1 \min \left\{1, \sum_{\mathcal{H} \in \mathbb{H}} \Pr_{D \sim Q} \left( L_D(\mathcal{H}) \leq \varepsilon_0  \;\middle|\; \lambda_Q(\mathcal{H}) \geq \gamma \right)^n \right\} d \gamma \\
        \intertext{Note that $\Pr_{D \sim Q} \left( L_D(\mathcal{H}) \leq \varepsilon_0 \right) = 1 - \lambda_Q(\mathcal{H})$. Therefore, for any choice of $p \in (0, 1)$:}
        &\leq \int_0^1 \min \left\{1, \sum_{\mathcal{H} \in \mathbb{H}} \left( 1 - \gamma \right)^n \right\} d \gamma
        \leq \int_0^p 1 d \gamma + \int_p^1 \lvert \mathbb{H} \rvert e^{-\gamma n} d \gamma \\
        &\leq p + \frac{\lvert \mathbb{H} \rvert}{n} e^{- p n}
    \end{align*}
    Now we are left with the second term. Using a union bound:
    \begin{align*}
        \mathds{1} \left( \exists \mathcal{H} \in \mathbb{H}[S], \exists i \in [n] : L_{D_i}(\mathcal{H}) > \varepsilon_0 \right)
        &\leq \sum_{\mathcal{H} \in \mathbb{H}} \mathds{1} \left( L_S(\mathcal{H}) = 0 \wedge \exists i \in [n] : L_{D_i}(\mathcal{H}) > \varepsilon_0 \right)
    \end{align*}
    taking the expectation over both $\mathbf{D} \sim Q^n$ and $S \sim \mathbf{D}^m$ we get:
    \begin{align*}
        \dE_{\mathbf{D} \sim Q^n} \dE_{S \sim \mathbf{D}^m} &\mathds{1} \left( \exists \mathcal{H} \in \mathbb{H}[S], \exists i \in [n] : L_{D_i}(\mathcal{H}) > \varepsilon_0 \right) \\
        &\leq \sum_{\mathcal{H} \in \mathbb{H}} \dE_{\mathbf{D} \sim Q^n} \dE_{S \sim \mathbf{D}^m} \mathds{1} \left( L_S(\mathcal{H}) = 0 \wedge \exists i \in [n] : L_{D_i}(\mathcal{H}) > \varepsilon_0 \right) \\
         &\leq \sum_{\mathcal{H} \in \mathbb{H}} \dE_{\mathbf{D} \sim Q^n} \Pr_{S \sim \mathbf{D}^m} \left( L_S(\mathcal{H}) = 0 \;\middle|\; \exists i \in [n] : L_{D_i}(\mathcal{H}) > \varepsilon_0 \right) \\
         \intertext{For each $\mathcal{H} \in \mathbb{H}[S]$ and $\mathbf{D} \in \mathcal{D}^n$, denote by $i_{\mathcal{H}, \mathbf{D}}$ the smallest index for which $L_{D_i}(\mathcal{H}) > \varepsilon_0$ if it exists, then}
         &\leq \sum_{\mathcal{H} \in \mathbb{H}} \dE_{\mathbf{D} \sim Q^n} \Pr_{S \sim \mathbf{D}^m} \left( L_{S_{i_{\mathcal{H}, \mathbf{D}}}}(\mathcal{H}) = 0 \;\middle|\; i_{\mathcal{H}, \mathbf{D}}, \mathbf{D} \right)
    \end{align*}
    Using standard $\varepsilon$-net bound, if $m \geq c \cdot \frac{1}{\varepsilon_0} \left(d \log \frac{1}{\varepsilon_0} + \log \frac{1}{\delta} \right)$ for some universal constant $c > 0$ then,
    \[
        \Pr_{S \sim \mathbf{D}^m} \left( L_{S_{i_{\mathcal{H}, \mathbf{D}}}}(\mathcal{H}) = 0 \;\middle|\; i_{\mathcal{H}, \mathbf{D}}, \mathbf{D} \right)
        \leq \delta
    \]
    Therefore,
    \begin{align*}
        \dE_{\substack{\mathbf{D} \sim Q^n \\ S \sim \mathbf{D}^m} } &\mathds{1} \left( \exists \mathcal{H} \in \mathbb{H}[S], \exists i \in [n] : L_{D_i}(\mathcal{H}) > \varepsilon_0 \right)
        \leq \lvert \mathbb{H} \rvert \cdot \delta
    \end{align*}
    Combining the bounds we have,
    \begin{align*}
        \dE_{S \sim Q^{(n, m)}} \sup_{\mathcal{H} : L_S(\mathcal{H})} L_Q(\mathcal{H})
        \leq \varepsilon_0 + p + \frac{\lvert \mathbb{H} \rvert}{n} e^{- p n} + \lvert \mathbb{H} \rvert \cdot \delta
    \end{align*}
    For all $p, \delta \in (0, 1)$ and $m \geq c \cdot \frac{1}{\varepsilon_0} \left(d \log \frac{1}{\varepsilon_0} + \log \frac{1}{\delta} \right)$. \\
    Finally, set $p = \frac{\log \lvert \mathbb{H} \rvert}{n}$ and $\delta = \frac{\varepsilon_0}{ \lvert \mathbb{H} \rvert}$.
    Then for all $m \geq 2c \cdot \frac{1}{\varepsilon_0} \left( d \log \frac{1}{\varepsilon_0} + \log \lvert \mathbb{H} \rvert \right)$ to get,
    \begin{align*}
        \dE_{S \sim Q^{(n, m)}} \sup_{\mathcal{H} : L_S(\mathcal{H})} L_Q(\mathcal{H})
        \leq 2 \varepsilon_0 + 2 \frac{\log \lvert \mathbb{H} \rvert}{n}
    \end{align*}
    Note that for $\varepsilon_0 = \frac{4c}{m} \left( d \log \frac{m}{d} + \log \lvert \mathbb{H} \rvert \right)$, the condition for $m$ is satisfied. Therefore,
    \[
        \dE_{S \sim Q^{(n, m)}} \sup_{\mathcal{H} : L_S(\mathcal{H})} L_Q(\mathcal{H})
        \leq 8c \frac{d \log \frac{m}{d} + \log \lvert \mathbb{H} \rvert}{m} + 2 \frac{\log \lvert \mathbb{H} \rvert}{n}
        = O_\mathbb{H} \left( \frac{1}{n} + \frac{\log m}{m} \right)
    \]
\end{proof}

\subsection{Number of domains}
We will prove a more general version of \Cref{Theorem: lower bound number of domains} for infinite meta-hypothesis families and all algorithms, not only ERMs.
We start by defining a property of the meta-hypothesis family inspired by game theory.
\begin{definition}[Value of meta-hypothesis family] \label{Definition: value of meta-hypothesis family}
    Let $\mathbb{H}$ be a meta-hypothesis family.
    Denote by $\Delta(\mathbb{H})$ the set of all distributions over the classes.
    We define the \underline{value of $\mathbb{H}$} as:
    \[
        v(\mathbb{H}) = \inf_{P \in \Delta(\mathbb{H})} \sup_{Q \in \text{RE}(\mathbb{H})} \dE_{\mathcal{H} \sim P} \dE_{D \sim Q} L_D(\mathcal{H})
    \]
\end{definition}
Intuitively, this can be thought of as a two-player zero-sum game between a naive algorithm and an adversary. The algorithm chooses a hypothesis class and the adversary chooses a domain, the utility of the adversary is the error of the chosen hypothesis class on the domain.

Note that a meta-hypothesis family with $v(\mathbb{H}) = 0$ is a trivial family in the sense that an algorithm could achieve arbitrary small without relying on any data. We will show this in the proof of the next result, \Cref{Theorem: lower bound number of domains - general case}.

Moreover, we define a simple assumption about the meta-hypothesis family that assumes the existence of a trivial task - a domain realizable by all classes:
\begin{definition}[Strong non-separability] \label{Assumption: strong non-separability}
    Let $\mathbb{H}$ be a meta-hypothesis family. We say that \underline{$\mathbb{H}$ satisfies the strong non-separability assumption} if there exists a meta-distribution $Q$ such that for all $\mathcal{H} \in \mathbb{H}$ we have $L_Q(\mathcal{H}) = 0$
\end{definition}
A simple case is when there is an example $(x, y) \in \mathcal{X} \times \mathcal{Y}$ that all classes can correctly classify.

The next theorem states that as long as there is a simple task, one realizable by all classes, and the family is non-trivial ($v(\mathbb{H}) > 0$) then the error of any proper algorithm is $\Omega_\mathbb{H} \left(\frac{1}{n} \right)$.
\begin{theorem}[General learning surface lower bound with respect to $n$] \label{Theorem: lower bound number of domains - general case}
    Let $\mathbb{H}$ be a meta-hypothesis family.
    Assume $\mathbb{H}$ satisfies \Cref{Assumption: strong non-separability}.
    \begin{itemize}
        \item If $v(\mathbb{H}) > 0$, then there exists $c > 0$ depending only on $\mathbb{H}$ such that for any proper learning algorithm $\mathcal{A}$ and any $n, m \in \mathbb{N}$:
        \[
            \sup_{Q \in \text{RE}(\mathbb{H})} \dE_{S \sim Q^{(n, m)}} L_Q(\mathcal{A}(S)) \geq \frac{c}{n}
        \]
        \item Otherwise, if $v(\mathbb{H}) = 0$, then for any rate $R\colon \mathbb{N} \times \mathbb{N} \to (0, 1]$ there exists an algorithm $\mathcal{A}$ with
        \[
        \sup_{Q \in \text{RE}(\mathbb{H})} \dE_{S \sim Q^{(n, m)}} L_Q(\mathcal{A}(S)) \leq R(n, m)
    \]
    \end{itemize}
\end{theorem}
\begin{proof}
    Assume $v(\mathbb{H}) > 0$.
    Let $\mathcal{A}$ be a learning algorithm and let $n, m \in \mathbb{N}$.
    Denote by $Q_0$ the meta-distribution that satisfies \Cref{Assumption: strong non-separability}.

    Define the distribution over $\mathbb{H}$ induced by the algorithm $P_\mathcal{A} = \dE_{S \sim Q_0^{(n, m)}} \mathcal{A}(S)$.
    % W.L.O.G the support of $Q_0$ and $Q_\mathcal{A}$ are disjoint.
    By definition of the value of $\mathbb{H}$, there exists some $Q_\mathcal{A} \in \text{RE}(\mathbb{H})$ such that
    $
        \dE_{\mathcal{H} \sim P_\mathcal{A}} L_{Q_\mathcal{A}}(\mathcal{H}) \geq \frac{v(\mathbb{H})}{2}
    $
    
    Next, define the set of realizable meta-distributions:
    $
        \left\{ Q_{\mathcal{A}, p} \coloneqq p \cdot Q_\mathcal{A} + (1 - p) \cdot Q_0 \;\middle|\; p \in (0, 1) \right\}
    $
    For any $p \in (0, 1)$:
    \begin{align*}
        \dE_{S \sim Q_{\mathcal{A}, p}^{(n, m)}} L_{Q_{\mathcal{A}, p}}(\mathcal{A}(S))
        &\geq p(1 - p)^n \dE_{S \sim Q_0^{(n, m)}} L_{Q_\mathcal{A}}(\mathcal{A}(S))
        = p(1 - p)^n \dE_{\mathcal{H} \sim P_\mathcal{A}} L_{Q_\mathcal{A}}(\mathcal{H}) \\
        &\geq p(1 - p)^n \frac{v(\mathbb{H})}{2} \\
        \intertext{Specifically, for $p = \frac{1}{n + 1}$}
        &= \frac{1}{n + 1} \left( 1 - \frac{1}{n + 1} \right)^n \frac{v(\mathbb{H})}{2}
        = \frac{v(\mathbb{H})}{2 n} \left(1 - \frac{1}{n + 1} \right)^{n + 1} \\
        &\geq \frac{v(\mathbb{H})}{8 n}
    \end{align*}
    Denote $c = \frac{v(\mathbb{H})}{8}$ to complete this part.

    Next, assume $v(\mathbb{H}) = 0$ and fix some $R \colon \mathbb{N} \times \mathbb{N} \to (0, 1]$.
    For any $n, m$, by the definition of $v(\mathbb{H})$ there exists a distribution $P_{n, m} \in \Delta(\mathbb{H})$ such that
    \[
        \sup_{Q \in \text{RE}(\mathbb{H})} \dE_{\mathcal{H} \sim P_{n, m}} L_Q(\mathcal{H}) \leq R(n, m)
    \]
    Define the algorithm $\mathcal{A}$ as follows:
    Given $S \in \left( \mathcal{X} \times \mathcal{Y} \right)^{(n, m)}$, sample $\mathcal{H} \sim P_{n, m}$ and output $\mathcal{H}$. Then,
    for any $Q \in \text{RE}(\mathbb{H})$,
    \[
        \dE_{S \sim Q^{(n, m)}} L_Q(\mathcal{A}(S))
        = \dE_{S \sim Q^{(n, m)}} \dE_{\mathcal{H} \sim P_{n, m}} L_Q(\mathcal{H})
        \leq \dE_{S \sim Q^{(n, m)}} R(n, m)
        = R(n, m)
    \]
    As needed.
\end{proof}

The proof of \Cref{Theorem: lower bound number of domains} follows a very similar proof using the assumptions of non-triviality together with the finiteness of the meta-hypothesis family.
\label{Proof: lower bound number of domains}
\begin{proof}[Proof of \Cref{Theorem: lower bound number of domains}]
    This proof follows similar ideas as in \Cref{Theorem: lower bound number of domains - general case}.
    We start by constructing a subset of $\mathbb{H}$ that satisfies \Cref{Assumption: strong non-separability} and has a positive value (as defined in \Cref{Definition: value of meta-hypothesis family}).
    By \Cref{Theorem: lower bound number of domains - general case}, any algorithm over this subset cannot have an error less than $\frac{c}{n}$ for all $n, m$.
    We then show that any algorithm over the entire $\mathbb{H}$ either outputs the subset with high enough probability or has a constant error.

    First, by \Cref{Assumption: weak non-separability}, there exists $(x_0, y_0) \in \mathcal{X} \times \mathcal{Y}$ such that $\lvert \mathbb{H}_0 \rvert \geq 2$ for
    \[
        \mathbb{H}_0 = \left\{ \mathcal{H} \in \mathbb{H} : \exists h \in \mathcal{H} \quad h(x_0) = y_0) \right\}
    \]
    Define the meta-distribution $Q_0 = \delta_{\delta_{(x_0, y_0)}}$ - a meta-distribution with mass 1 on the domain which has mass 1 on $(x_0, y_0)$.
    Note that for any $\mathcal{H} \in \mathbb{H}_0$ we have $L_{Q_0}(\mathcal{H}) = 0$. Thus, $\mathbb{H}_0$ satisfy \Cref{Assumption: strong non-separability} with $Q_0$.

    We will now show that $v(\mathbb{H}_0) > 0$.
    Using \Cref{Assumption: no pairwise domination}, for any $\mathcal{H} \in \mathbb{H}_0$ there is $\mathcal{H}' \in \mathbb{H}_0$ and a meta-distribution $Q_\mathcal{H}$ such that
    \[
        L_{Q_\mathcal{H}}(\mathcal{H}) > 0 \quad L_{Q_\mathcal{H}}(\mathcal{H}') = 0
    \]
    Denote $c_0 = \min_{\mathcal{H} \in \mathbb{H}_0} L_{Q_\mathcal{H}}(\mathcal{H})$, and note that $c_0 > 0$.
    For any $P \in \Delta(\mathbb{H}_0)$, there exists $\mathcal{H} \in \mathbb{H}_0$ with $P(\mathcal{H}) \geq \frac{1}{\lvert \mathbb{H}_0 \rvert} > 0$. Therefore,
    \[
        \dE_{\mathcal{H}' \sim P} L_{Q_\mathcal{H}}(\mathcal{H}')
        \geq P(\mathcal{H}) \cdot L_{Q_\mathcal{H}}(\mathcal{H})
        \geq \frac{c_0}{\lvert \mathbb{H}_0 \rvert}
        > 0
    \]
    This implies that $v(\mathbb{H}_0) > 0$.

    Let $\mathcal{A}$ be a learning algorithm. Similar to the proof of \Cref{Theorem: lower bound number of domains - general case}, define the induced distribution $P_\mathcal{A} = \dE_{S \sim Q_0^{(n, m)}} \mathcal{A}(S)$.
    
    We differentiate between two cases:
    
    \underline{Case 1} - $\Pr_{\mathcal{H} \sim P_\mathcal{A}}(\mathcal{H} \in \mathbb{H}_0) < \frac{1}{2}$:
    Since for all $\mathbb{H} \notin \mathbb{H}_0$ we have $L_{Q_0}(\mathcal{H}) = 1$ we get
    \[
        \dE_{S \sim Q_0^{(n, m)}} L_{Q_0}(\mathcal{A}(S))
        \geq \frac{1}{2}
    \]
    So the error of the algorithm is bounded by a constant.

    \underline{Case 2} - $\Pr_{\mathcal{H} \sim P_\mathcal{A}}(\mathcal{H} \in \mathbb{H}_0) \geq \frac{1}{2}$:
    Define the conditional distribution induced by the algorithm $P_{\mathcal{A}, 0} = \dE_{S \sim Q^{(n, m)}} \left[ \mathcal{A}(S) \;\middle|\; \mathcal{A}(S) \in \mathbb{H}_0 \right]$.
    Note that $P_{\mathcal{A}, 0} \in \Delta(\mathbb{H}_0)$.
    Therefore, there exists a meta-distribution $Q_{\mathcal{A}, 0} \in \text{RE}(\mathbb{H}_0)$ such that $\dE_{\mathcal{H} \sim P_{\mathcal{A}, 0}} L_{Q_{\mathcal{A}, 0}}(\mathcal{H}) \geq \frac{v(\mathbb{H}_0)}{2}$.
    
    Following the proof of \Cref{Theorem: lower bound number of domains - general case}, using the meta-distribution $Q_{\mathcal{A}, p} = p \cdot Q_{\mathcal{A}, 0} + (1 - p) Q_0$:
    \[
        \dE_{S \sim Q_{\mathcal{A}, p}^{(n, m)}} L_{Q_{\mathcal{A}, p}}
        \geq p (1 - p)^n \frac{v(\mathbb{H}_0)}{4}
    \]
    Setting $p = \frac{1}{n + 1}$ to get
    \[
        \dE_{S \sim Q_{\mathcal{A}, p}^{(n, m)}} L_{Q_{\mathcal{A}, p}}
        \geq \frac{v(\mathbb{H}_0)}{8 n}
    \]
    For $c = \frac{v(\mathbb{H}_0)}{8}$ we complete the proof.
\end{proof}

\subsection{Number of examples per domain}
\begin{proof}[Proof of \Cref{Theorem: bound number of examples per domain}] \label{Proof: bound number of examples per domain}
    We prove the theorem with two inequalities.
    We start with the upper bound:
    $\varepsilon^\text{ERM}_\text{exp}(m) \leq \varepsilon_\mathbb{H}(m)$

    Let $m \in \mathbb{N}$, and fix $\varepsilon > \varepsilon_\mathbb{H}(m)$.
    For any meta-distribution $Q \in \text{RE}(\mathbb{H})$ and any class $\mathcal{H} \in \mathbb{H}$ with $L_Q(\mathcal{H}) \geq \varepsilon$.
    We will show that any ERM would choose such a hypothesis class with probability zero (as $n \to \infty$).

    For this, we suggest an alternative sampling method:
    \begin{enumerate}
        \item Sample $\mathbf{D} \sim Q^n$ as usual.
        \item Sample a large set $S' \sim \mathbf{D}^T$ with $T = O \left( \frac{1}{\varepsilon^2} \left(d + \log \frac{1}{\delta} \right) \right)$
        This set will have a non-realizable set with high probability.
        \item Resample from the large set $S \sim S'^{(m)}$ which denotes sampling with no repetitions.
    \end{enumerate}
    Note that the distribution of $S$ is equivalent to our usual sampling method $S \sim Q^{(n, m)}$.

    We will further denote several events. Each one will be a possible error source that corresponds to a different stage in the alternative sampling process:
    \begin{align*}
        E_1 &= \lvert \left\{ i : L_{D_i} (\mathcal{H}) \geq \beta \varepsilon \right\} \rvert < (1 - \alpha) (1 - \beta) n \varepsilon \\
        E_2 &= \lvert \left\{ i : L_{S'_i} (\mathcal{H}) \geq \beta \gamma \varepsilon \right\} \rvert < (1 - \alpha) (1 - \beta) (1 - \eta) (1 - \delta) n \varepsilon \\
        E_3 &= \lvert \left\{ i : L_{S_i} (\mathcal{H}) \geq 0 \right\} \rvert = 0
    \end{align*}
    With $\alpha, \beta. \gamma, \delta, \eta \in (0, 1)$ to be specified later.

    Using conditioning, we have:
    \begin{align*}
        \Pr_{S \sim Q^{(n, m)}} \left( L_S(\mathcal{H}) = 0 \right)
        \leq \Pr_{\mathbf{D} \sim Q^n} \left(E_1 \right)
        + \Pr_{\substack{\mathbf{D} \sim Q^n \\ S' \sim \mathbf{D}^T}} \left( E_2 \mid \neg E_1 \right)
        + \Pr_{\substack{\mathbf{D} \sim Q^n \\ S' \sim \mathbf{D}^T \\ S \sim S'^{(m)}}} \left( E_3 \mid \neg E_1 \enspace \neg E_2 \right)
    \end{align*}
    We bound each term individually.

    For the first term, using Markov's inequality:
    \begin{align*}
        \Pr_{D \sim Q} \left( L_D(\mathcal{H}) < \beta \varepsilon \right)
        &= \Pr_{D \sim Q} \left( 1 - L_D(\mathcal{H}) > 1 - \beta \varepsilon \right)
        \leq \frac{1 - \dE_{D \sim Q} L_D(\mathcal{H})}{1 - \beta \varepsilon} \\
        &\leq \frac{1 - \varepsilon}{1 - \beta \varepsilon}
        = 1 - \frac{(1 - \beta) \varepsilon}{1 - \beta \varepsilon}
        \leq 1 - (1 - \beta) \varepsilon
    \end{align*}
    And therefore,
    \[
        \dE_{\mathbf{D} \sim Q^n} \sum_{i=1}^{n} \mathds{1} \left(L_{D_i}(\mathcal{H}) \geq \beta \varepsilon \right)
        \geq (1 - \beta) n \varepsilon
    \]
    Using (multiplicative) Chernoff bound:
    \begin{align*}
        \Pr_{\mathbf{D} \sim Q^n} \left( E_1 \right)
        &\leq \Pr_{\mathbf{D} \sim Q^n} \left( \sum_{i=1}^{n} \mathds{1} \left(L_{D_i}(\mathcal{H}) \geq \beta \varepsilon \right) \leq (1 - \alpha) \dE\left[ \sum_{i=1}^{n} \mathds{1} \left(L_{D_i}(\mathcal{H}) \geq \beta \varepsilon \right) \right] \right) \\
        &\leq \exp \left( - \frac{1}{2} \alpha^2 \dE \left[\sum_{i=1}^{n} \mathds{1} \left(L_{D_i}(\mathcal{H}) \geq \beta \varepsilon \right) \right] \right)
        \leq \exp \left( - \frac{1}{2} \alpha^2 (1 - \beta) n \varepsilon \right)
    \end{align*}

    For the second term, using uniform convergence results (for agnostic learning), for $T = O \left( \frac{d + \log \frac{1}{\delta}}{(\beta (1 - \gamma) \varepsilon)^2} \right)$ we have:
    \[
        \Pr_{S' \sim D^T} \left( \lvert L_{S"}(\mathcal{H}) -L_D(\mathcal{H}) \rvert \leq \beta (1 - \gamma) \varepsilon \right)
        \geq 1 - \delta
    \]
    For any domain with $L_D(\mathcal{H}) \geq \beta \varepsilon$ this implies:
    \begin{align*}
        \Pr_{S' \sim \mathbf{D}^T} \left( L_{S'}(\mathcal{H}) \geq \beta \gamma \varepsilon \right)
        &\geq \Pr_{S' \sim \mathbf{D}^T} \left( L_{S'}(\mathcal{H}) \geq L_D(\mathcal{H}) - \beta (1 - \gamma) \varepsilon \right) \\
        &\geq \Pr_{S' \sim D^T} \left( \lvert L_{S"}(\mathcal{H}) -L_D(\mathcal{H}) \rvert \leq \beta (1 - \gamma) \varepsilon \right)
        \geq 1 - \delta
    \end{align*}
    Therefore, 
    \[
        \dE_{\substack{\mathbf{D} \sim Q^n \\ S' \sim \mathbf{D}^T}} \left[ \sum_{i=1}^{n} \mathds{1} \left(L_{S'_i}(\mathcal{H}) \geq \beta \gamma \varepsilon \right) \;\middle|\; \neg E_1 \right]
        \geq (1 - \delta) (1 - \alpha) (1 - \beta) n \varepsilon
    \]
    And using Chernoff bound once again we have:
    \begin{align*}
        \Pr_{\substack{\mathbf{D} \sim Q^n \\ S' \sim \mathbf{D}^T}} \left( E_1 \;\middle|\; \neg E_2 \right)
        \leq \exp \left( - \frac{1}{2} \eta^2 (1 - \alpha) (1 - \beta) (1 - \delta) n \varepsilon \right)
    \end{align*}

    For the third term, for any $\beta, \gamma \in (0, 1)$ with $\beta \gamma \varepsilon > \varepsilon_\mathbb{H}(m)$ we have $m' \coloneqq m_\mathbb{H}(\beta \gamma \varepsilon) \leq m$.
    Thereby, for any $S' \subseteq \mathcal{X} \times \mathcal{Y}$ with $L_{S'}(\mathbb{H}) = 0$ and $L_{S'}(\mathcal{H}) \geq \beta \gamma \varepsilon$ there exists some $S \subseteq S'$ with $\lvert S' \rvert \leq m'$ and $L_S(\mathcal{H}) > 0$.
    Sampling a subset of size $m$ from $S'$ uniformly will contain this non-realizable subset with a probability of at least
    $
        \frac{\binom{T - m'}{m - m'}}{\binom{T}{m}}
        \geq \frac{1}{\binom{T}{m}}
    $.
    This implies
    \[
        \Pr_{\substack{D \sim Q \\ S' \sim D^T \\ S \sim S'^{(m)}}} \left( L_S(\mathcal{H}) = 0 \;\middle|\; L_{S'}(\mathcal{H}) \geq \beta \gamma \varepsilon \right)
        \leq 1 - \binom{T}{m}^{-1}
        \leq \exp \left( - \binom{T}{m}^{-1} \right)
    \]
    Conditioning on the negation of the two previous events we get:
    \[
        \Pr_{\substack{D \sim Q \\ S' \sim D^T \\ S \sim S'^{(m)}}} \left( E_3 \;\middle|\; \neg E_1 \enspace \neg E_2 \right)
        \leq \exp \left( - (1 - \alpha) (1 - \beta) (1 - \eta) (1 - \delta) \binom{T}{m}^{-1} n \varepsilon \right)
    \]
    Combining the three terms:
    \begin{align*}
        \Pr_{S \sim Q^{(n, m)}} \left( L_S(\mathcal{H}) = 0 \right)
        &\leq \exp \left( - \frac{1}{2} \alpha^2 (1 - \beta) n \varepsilon \right) \\
        &+ \exp \left( - \frac{1}{2} (1 - \alpha) (1 - \beta) (1 - \delta) \eta^2 n \varepsilon \right) \\
        &+ \exp \left( - (1 - \alpha) (1 - \beta) (1 - \delta) (1 - \eta) \binom{T}{m}^{-1} n \varepsilon \right)
    \end{align*}
    Note this bound is uniform for all $\mathcal{H} \in \mathbb{H}$ and $Q \in \text{RE}(\mathbb{H})$ with $L_Q(\mathcal{H}) \geq \varepsilon$.
    Using a union bound over $\mathbb{H}$:
    \[
        \Pr_{S \sim Q^{(n, m)}} \left( \sup_{\mathcal{H} : L_S(\mathcal{H}) = 0} L_Q(\mathcal{H}) \geq \varepsilon \right)
        \leq \sum_{\mathcal{H} : L_S(\mathcal{H}) = 0} \Pr_{S \sim Q^{(n, m)}} \left( L_S(\mathcal{H}) = 0 \right)
    \]
    This tends to $0$ as $n \to \infty$.
    Finally, since
    \[
        \dE_{S \sim Q^{(n, m)}} \sup_{\mathcal{H} : L_S(\mathcal{H}) = 0} L_Q(\mathcal{H}) \leq \varepsilon + \Pr_{S \sim Q^{(n, m)}} \left( \sup_{\mathcal{H} : L_S(\mathcal{H}) = 0} L_Q(\mathcal{H}) \geq \varepsilon \right)
    \]
    we have $\varepsilon^\text{ERM}_\text{exp}(m) \leq \varepsilon$ for all $\varepsilon > \varepsilon_\mathbb{H}(m)$.

    For the lower bound $\varepsilon^\text{ERM}_\text{exp}(m) \geq \varepsilon_\mathbb{H}(m)$, fix $m \in \mathbb{N}$ and fix some $0 < \varepsilon < \varepsilon_\mathbb{H}(m)$. (The case for $\varepsilon = 0$ is trivial).

    By \Cref{Definition: optimal error function}, $m_\mathbb{H}(\varepsilon) > m$ and therefore, there exists $\mathcal{H} \in \mathbb{H}$ with $m_{\mathcal{H} \mid \mathbb{H}}(\varepsilon) > m$.

    By the negation of the $\varepsilon$ dual Helly number, there exists $S^* \subseteq \mathcal{X} \times \mathcal{Y}$ with $L_{S^*}(\mathbb{H}) = 0$ and $L_{S^*}(\mathcal{H}) > \varepsilon$ yet for all $S \subseteq S^*$ with $\lvert S \rvert \leq m$ we have $L_S(\mathcal{H}) = 0$.

    For the domain which samples from $S^*$ uniformly, $D = U(S^*)$ and the meta-distribution that always outputs this domain, $Q = \delta_D$, any sampled sets $S \sim Q^{(n, m)}$ is realizable by $\mathcal{H}$ with probability 1, yet $L_Q(\mathcal{H}) > \varepsilon$.

    This implies that for all $n$:
    \[
        \dE_{S \sim Q^{(n, m)}} \sup_{\mathcal{H} : L_S(\mathcal{H}) = 0} L_Q(\mathcal{H}) > \varepsilon
    \]
    And therefore, $\varepsilon^\text{ERM}_\text{exp}(m) \geq \varepsilon$ for all $\varepsilon < \varepsilon_\mathbb{H}(m)$
\end{proof}

We will next prove a lower bound on the rate of $\varepsilon_\mathbb{H}$ in the case where the dual Helly number is infinite:
\begin{proof}[Proof of \Cref{Lemma: Rates of dual Helly}] \label{Proof: rates of dual Helly}
    First, assume $m_\mathbb{H}(0) < \infty$.
    For all $\varepsilon > 0$ we have $m_\mathbb{H}(\varepsilon) \leq m_\mathbb{H}(0)$.
    Therefore, for all $m \geq m_\mathbb{H}(0)$ we have $\varepsilon_\mathbb{H}(m) = 0$ by definition of the optimal error function.
    
    Next, assume $m_\mathbb{H}(0) = \infty$. Then, for any $k \in \mathbb{N}$ there exists $\mathcal{H} \in \mathbb{H} \in \mathbb{H}$ with $m_\mathcal{H}(0) > k$.
    From the definition of $m_\mathbb{H}(0)$, there exists $S \subseteq \mathcal{X} \times \mathcal{Y}$ with $L_S(\mathcal{H}) > 0$ yet $L_S(\mathbb{H}) = 0$ and for any subset $S' \subseteq S$ with $\lvert S' \rvert < k$ we have $L_{S'}(\mathcal{H}) = 0$

    Note that for some $k' \geq k$, there exists $S^* \subseteq S$ with $\lvert S^* \rvert = k'$ with $L_{S*}(\mathcal{H}) > 0$. Denote by $k^*$ the minimal such size and note that $k^* \geq k$.
    Then, $L_{S^*}(\mathcal{H}) > 0 \implies L_{S^*}(\mathcal{H}) \geq \frac{1}{k^*}$ and by its minimality, all subsets $S' \subseteq S^*$ satisfy $L_{S'}(\mathcal{H}) = 0$.
    Next, define the meta-distribution $Q = \delta_{U(S^*)}$ which samples from $S^*$ uniformly. For all $m < k^*$, any sampled set $S$ of size $m$ is a (multi-)subset of $S^*$ and therefore $L_S(\mathcal{H}) = 0$ yet $L_Q(\mathcal{H}) = L_{S^*}(\mathcal{H}) \geq \frac{1}{k^*}$. Specifically, for $m = k^* - 1$, $L_Q(\mathcal{H}) \geq \frac{1}{m + 1}$.

    Next, denote this $k^*$ as $k^*_1$, and repeat the argument with $k = k^*_1 + 1$. This gives a new minimal $k^* > k^*_1$ with a corresponding classes $\mathcal{H}$ and sets $S^*$ with $L_{S^*}(\mathcal{H}) > 0$. Denote the new $k^*$ as $k^*_2$ and repeat to create a series $\{k^*_i\}_{i \in \mathbb{N}}$.
    Denote their corresponding hypothesis classes as $\mathcal{H}_i$ and meta-distributions as $Q_i$.

    Finally, define the ERM algorithm $\mathcal{A}$ such that if $m = k^*_i - 1$ for some $i$, then it outputs $\mathcal{H}_i$. By definition of our series, for any such $m$,
    \begin{align*}
        \varepsilon_\mathbb{H}(m)
        &\geq \sup_{Q \in \text{RE}(\mathbb{H})} \dE_{S \sim Q^m} L_Q(\mathcal{A}(S))
        \geq \dE_{S \sim Q_i^m} L_{Q_i}(\mathcal{A}(S))
        \geq \frac{1}{m + 1}
    \end{align*}
\end{proof}

\subsection{Independence result}
To show the meta-learnability of $\mathbb{H}^*$ is independent from the ZFC axioms we will show that $\mathbb{H}^*$ is learnable if and only if the cardinality of the continuum is $\aleph_k$ for some $k \in \mathbb{N}$ which is known to be independent of the ZFC axioms. For more discussion about this independence result see \citet{ben2019learnability} and citations within.

\begin{proof}[Proof of \Cref{Theorem: independence result}] \label{Proof: independence result}
    We first show that if there exists a meta-algorithm $\mathcal{A}$ that properly meta-learns $\mathbb{H}^*$, then the family $\mathcal{F}^* = \left\{f_X(x) = \mathds{1}(x \in X) \mid X \subseteq [0, 1], \enspace \lvert X \rvert < \infty \right\}$ is (\nicefrac{1}{3}, \nicefrac{1}{3})-EMX learnable. According to \citet{ben2019learnability} this shows the cardinality of the continuum is $\aleph_k$ for some $k \in \mathbb{N}$. 

    Let $P$ be a distribution over $X \subseteq [0, 1]$ with $\lvert X \rvert < \infty$.
    For each $x \in X$, define the domain $D_x = \delta_x$ and the meta-distribution $Q$ such that $\Pr_{D \sim Q}(D = D_x) = P(x)$.

    Since $\mathcal{A}$ properly meta-learn $\mathbb{H}^*$, there exists some $n_0, m_0$ such that
    $\dE_{S \sim Q^{(n_0, m_0)}} L_Q(\mathcal{A}(S)) \leq \frac{1}{9}$.
    
    Define the algorithm $G \colon \mathcal{X}^{n_0} \to \mathcal{F}^*$ as follows:
    Given a sample $S'$ of $n_0$ examples from $X$, assign $y = 1$ for all and duplicate each example $m_0$ times to get $S \in (\mathcal{X} \times \mathcal{Y})^{(n, m)}$.
    Then, denote $\mathcal{H} = \mathcal{A}(S)$ and return the function $f_X \in \mathcal{F}^*$ for $X = \left\{ x \in \mathcal{X} \;\middle|\; \mathds{1}_{\{x\}} \in \mathcal{H} \right\}$.
    Then, using Markov's inequality,
    \begin{align*}
        \Pr_{S' \sim P^{n_0}} \left( \dE_P G(S') < \frac{2}{3} \right)
        = \Pr_{S \sim Q^{(n_0, m_0)}} \left( L_Q(\mathcal{A}(S)) > \frac{1}{3} \right)
        \leq \frac{\dE \left[L_Q(\mathcal{A}(S)) \right]}{\nicefrac{1}{3}}
        \leq \frac{1}{3}
    \end{align*}
    Which, by definition, means that $G$ is a $(\frac{1}{3}, \frac{1}{3})$-EMX learner for $\mathcal{F}^*$

    To show the other direction, we need to show that given the cardinality of the continuum is $\aleph_k$ for $k \in \mathbb{N}$ there is an algorithm $\mathcal{A}$ that meta-learn $\mathbb{H}^*$.
    Using the results from \citet{ben2019learnability}, if the cardinality of the continuum is $\aleph_k$ for $k \in \mathbb{N}$ then there exists a monotone compression scheme $n \to k + 1$ for some $k \in \mathbb{N}$  and all $n > k +1$ for $\mathcal{F}^*$. Denote $p = k + 1$ for simplicity.

    The compression scheme is defined using a reconstruction function $\eta \colon \mathcal{X}^p \to \mathcal{F}^*$ such that for every $f \in \mathcal{F}^*$ and $w_1, \dots, w_n \in \mathcal{X}$ with $f(w_i) = 1$ for all $i$, there exists $i_1, \dots, i_p$ so that for $g = \eta( (w_{i_1}, \dots, w_{i_p}) )$ we have $g(w_i) = 1$ for all $i$.
    
    We will define the algorithm $\mathcal{A}$ that learns using $m = 1$ examples per domain as follows:
    Let $S \in (\mathcal{X} \times \mathcal{Y})^{(n, 1)}$. For each $i \in [n]$, denote $S_i = \{(x_i, y_i)\}$. If $y_{i, j} = 1$, set $w_i = x_i$, otherwise set $w_i = 0$.
    Next, apply the reconstruction function $g = \eta( (w_{i_1}, \dots, w_{i_p}) )$ and return $\mathcal{H}_W$ for $W = \left\{w \in \mathcal{X} \;\middle|\; g(w) = 1\right\}$.

    We are left to show that $\mathcal{A}$ meta-learns $\mathbb{H}$.
    For every subset $B \subseteq [n]$ with $\lvert B \rvert = p$, let $\mathcal{H}_B$ denote the hypothesis class corresponding to the function $\eta((w_i)_{i \in B})$. Notice that $\mathcal{H}_B$ is independent of all domains $i \notin B$ and there are $n - p$ such domains.
    Note that $L_Q(\mathcal{A}(S)) \geq \varepsilon$ implies that at least one of $\mathcal{H}_B$ satisfies $L_Q(\mathcal{H}_B) \geq \varepsilon$ yet $L_S(\mathcal{H}_B) = 0$.

    For any fixed $B$,
    \[
        L_Q(\mathcal{H}_B) = \Pr_{\substack{D \sim Q \\ (x, y) \sim D}} \left( y = 1, \mathds{1}_{\{x\}} \notin \mathcal{H}_B \right)
    \]
    Conditioning on $L_Q(\mathcal{H}_B) \geq \varepsilon$,
    \[
        \Pr_{S \sim Q^{(n, 1)}} \left( L_S(\mathcal{H}_B) = 0 \;\middle|\; B \right)
        \leq \prod_{i \notin B} \Pr_{\substack{D \sim Q \\ (x, y) \sim D}} \left( y = 1, \mathds{1}_{\{x\}} \notin \mathcal{H}_B \right)
        \leq \left( 1 - \varepsilon \right)^{n - p}
    \]
    Using a union bound over all $B \subseteq [n]$ with $\lvert B \rvert \leq p$ we have,
    \[
        \Pr_{S \sim Q^{(n, 1)}} \left( L_Q(\mathcal{A}(S)) \geq \varepsilon \right)
        \leq \sum_{B} \Pr_{S \sim Q^{(n, 1)}} \left( L_S(\mathcal{H}_B) = 0, L_Q(\mathcal{H}_B \geq \varepsilon \;\middle|\; B \right)
        \leq \binom{n}{p} (1 - \varepsilon)^{n - p}
    \]
    This tends to zero as $n \to \infty$, and therefore the algorithm successfully meta-learns $\mathbb{H}^*$
\end{proof}

\end{document}